\documentclass[english]{lni}

\usepackage{url}
\usepackage{graphicx}
\usepackage{fancyhdr}
\usepackage{listings} 
\usepackage{changepage} 
\usepackage[figurename=Fig., tablename=Tab., small]{caption}[2008/04/01]

\usepackage{subcaption}

\fancypagestyle{titlepage}{
\fancyhead[RO]{\small N. Damer,  M. Gomez-Barrero,  K. Raja,  C. Rathgeb,  A.  Sequeira,\linebreak  M. Todisco,  and A. Uhl (Eds.): BIOSIG 2023, \linebreak Lecture Notes in Informatics (LNI), Gesellschaft f\"ur Informatik, Bonn 2023} 
\fancyfoot{}}

\renewcommand{\headrulewidth}{0.4pt} 
\author{Wassim Kabbani \footnote{IIK, Info. Sec. and Comm. Technology, Gjovik, Norway, wassim.h.kabbani@ntnu.no} \ , Christoph Busch \footnote{IIK, Info. Sec. and Comm. Technology, Gjovik, Norway, christoph.busch@ntnu.no} \ , Kiran Raja \footnote{IIKiran RajaKiran Raja, Info. Sec. and Comm. Technology, Gjovik, Norway, kiran.raja@ntnu.no}}
\title{Robust Sclera Segmentation for Skin-tone Agnostic Face Image Quality Assessment}
\begin{document}

\maketitle

\renewcommand{\refname}{References}
\setcounter{footnote}{3} 
\thispagestyle{titlepage}
\pagestyle{fancy}
\fancyhead{} 
\fancyhead[RO]{\small Robust Sclera Segmentation for Skin-tone Agnostic Face Image Quality Assessment \hspace{25pt}  \hspace{0.05cm}}
\fancyhead[LE]{\hspace{0.05cm}\small  \hspace{25pt} Wassim Kabbani and Christoph Busch and Kiran Raja}
\fancyfoot{} 
\renewcommand{\headrulewidth}{0.4pt} 

\begin{abstract}
Face image quality assessment (FIQA) is crucial for obtaining good face recognition performance. FIQA algorithms should be robust and insensitive to demographic factors. The eye sclera has a consistent whitish color in all humans regardless of their age, ethnicity and skin-tone. This work proposes a robust sclera segmentation method that is suitable for face images in the enrolment and the border control face recognition scenarios. It shows how the statistical analysis of the sclera pixels produces features that are invariant to skin-tone, age and ethnicity and thus can be incorporated into FIQA algorithms to make them agnostic to demographic factors.
\end{abstract}
\begin{keywords}
FIQA, Face Recognition, Facial Landmarks, Eye Sclera, Skin-tone, Illumination, Natural Color, Color Imbalance
\end{keywords}

\section{Introduction}

Face image quality assessment refers to the process of evaluating the utility of a face image for face recognition. It involves analyzing various quality factors that may impact the recognition performance. The quality measures produced from analyzing the image can be in the form of individual quality components, such as background uniformity, illumination uniformity, pose, exposure, dynamic range, sharpness, facial expressions, or in the form of a unified quality score.

The ISO/IEC CD on 29794-5 \cite{29794-5} (Information technology — Biometric sample quality — Part 5: Face image data) specifies that a face image quality assessment algorithm should be insensitive to demographic factors such as age, skin-tone or ethnicity.

\begin{figure}
     \centering
     \begin{subfigure}[b]{0.40\textwidth}
         \centering
         \includegraphics[width=\textwidth]{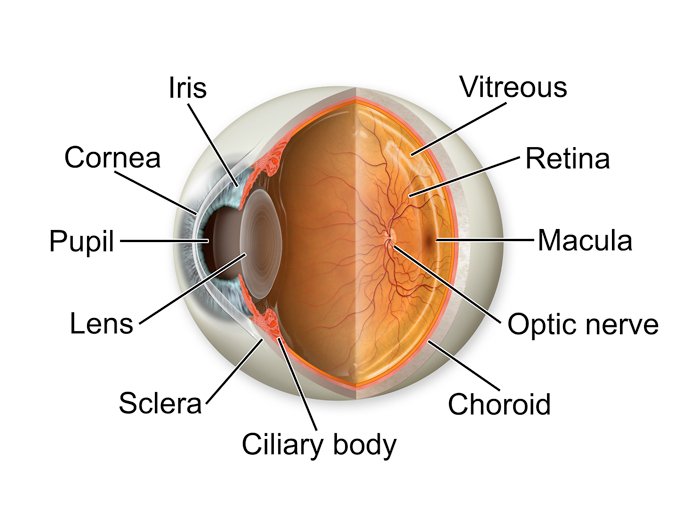}
         \caption{}
         \label{fig:ae-1}
     \end{subfigure}
     \begin{subfigure}[b]{0.40\textwidth}
         \centering
         \includegraphics[width=\textwidth]{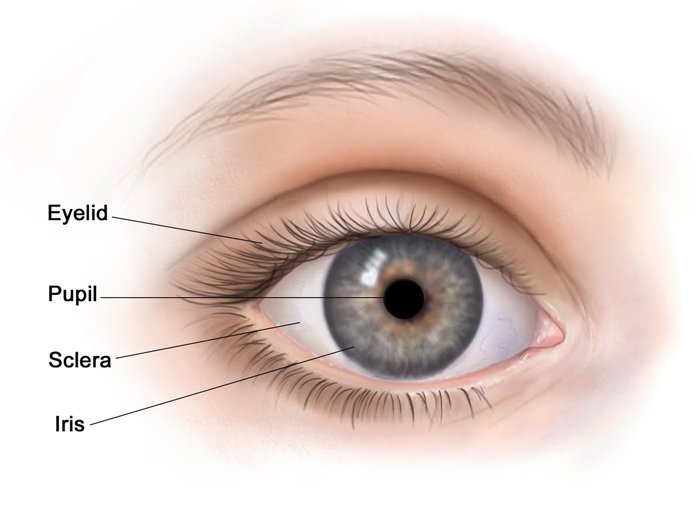}
         \caption{}
         \label{fig:ae2}
     \end{subfigure}
    \caption{Anatomy of the Eye. Image src: perspectiveopticians.co.uk}
    \label{fig:ae}
\end{figure}

The eye sclera refers to the outer layer of the eyeball surrounding the iris. It is the opaque, whitish portion of the eye that surrounds the colored iris and the dark circular opening called the pupil. Figure \ref{fig:ae} illustrates the anatomy of the eye including the sclera. This characteristic of being whitish in color regardless of age, ethnicity and skin-tone \cite{KANO202310} is what makes it interesting for the task of face image quality assessment.

Analyzing the eye sclera in a face image can help in making the quality assessment algorithms of some of the face image quality components invariant to skin-tone and ethnicity. Not all of the quality components specified in ISO/IEC CD on 29794-5 can make use of this but many of them can. This can be, for example, useful for all the illumination and color related components such as illumination uniformity, no under or over exposure, natural color, dynamic range as well as for eyes visible and eyes open.

\section{Related Work}

The blood vessels structure of the eye sclera is unique to each person, hence it could be used for identification \cite{sclreg}. Therefor, sclera segmentation methods have started to emerge since a high quality segmentation of the sclera from the eye and the iris is required before any further processing and recognition can take place.

Most sclera segmentation methods are deep learning based models trained on large scale datasets with ground truth segmentation masks. Some of them can only segment the sclera, others perform full eye segmentation for the sclera, the iris, and the pupil.

ScleraSegNet \cite{wang2019sclerasegnet} is a sclera segmentation method based on an attention assisted U-Net model. It utilizes attention modules in addition to the U-Net to imporve the segmentation performance. In a following improved version of ScleraSegNet \cite{wang2020sclerasegnet}, the authors suggest to adjust the architecture by placing the attention modules into the central bottelneck part between the contracting path and the expansive path of the U-Net to strengthen the learning capacity of the network and this proves to improve the segmentation performance.

RITnet \cite{chaudhary2019ritnet} is a real-time eye segmentation deep neural network model that is trained on the OpenEDS dataset \cite{OpenEDS}. It is the winning model of OpenEDS Semantic Segmentation Challenge 2019 \footnote{https://research.facebook.com/openeds-challenge/} and achieves state-of-the-art results on the OpenEDS's testset.

Segmentation models are usually trained on specialized datasets collected for training eye segmentation models and gaze tracking models. Among the largest and most recent datasets are OpenEDS \cite{OpenEDS}, and NVGaze \cite{NVGaze}.

The OpenEDS:OpenEyeDataset \cite{OpenEDS}, contains eye-images captured using a virtual-reality head mounted display with two eye-facing cameras and under controlled illumination. It contains high quality images of 400x640 pixels of the eye region only.

The NVGaze \cite{NVGaze} dataset, from Nvidia, is created to satisfy the criteria for near-eyegaze estimation under infrared illumination. It comprised two types of images, synthetic images of 1280x960 pixels, and real images of 640x480 pixels of the eye region only.

While face images could be of non-uniform illumination, imbalanced color, and low resolution, existing eye segmentation models are trained on datasets of high-resolution images captured under controlled illumination in a specialized setting. This makes them less suited for the task of segmenting the eye region in a face image in order to perform further analysis. Furthermore, the face parsing network \footnote{https://github.com/zllrunning/face-parsing.PyTorch} that is standardized in ISO/IEC CD 29794-5 \cite{29794-5} and which segments the face into 19 classes such as hair, eyeglasses, eyes, eyebrows, nose, mouth and ears, does not give a segmentation for the different regions inside the eye but rather for the eye as a whole. Thus, a dedicated sclera segmentation method that is suitable for face images commonly encountered during the face recognition process is needed.

\begin{figure}
     \centering
     \begin{subfigure}[b]{0.45\textwidth}
         \centering
         \includegraphics[width=\textwidth]{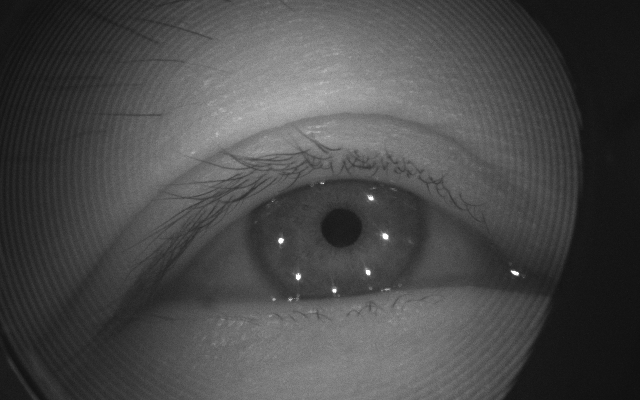}
         \includegraphics[width=\textwidth]{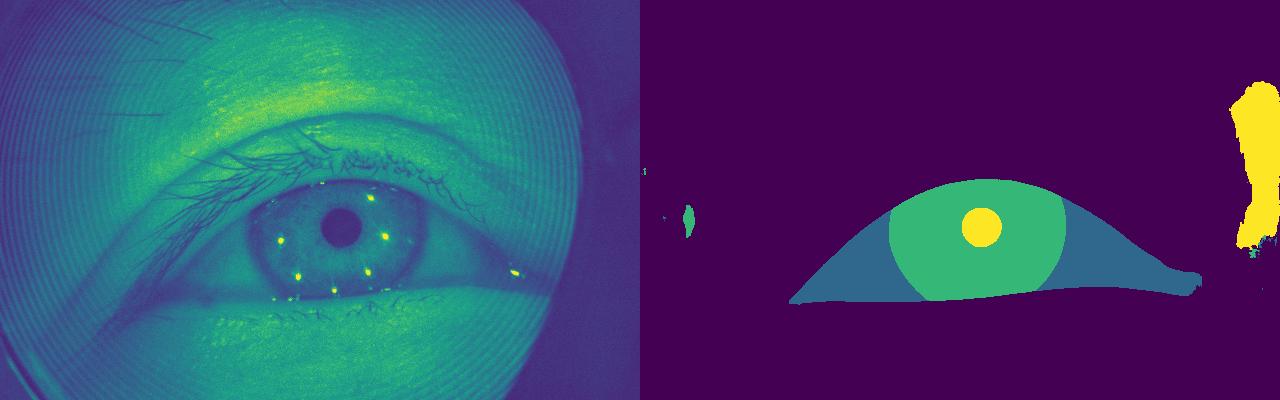}
         \caption{Image from OpenEDS \cite{OpenEDS}}
         \label{fig:seg-ritnet-1}
     \end{subfigure}
     \hfill
     \hfill
     \begin{subfigure}[b]{0.45\textwidth}
         \centering
         \includegraphics[width=\textwidth]{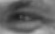}
         \includegraphics[width=\textwidth]{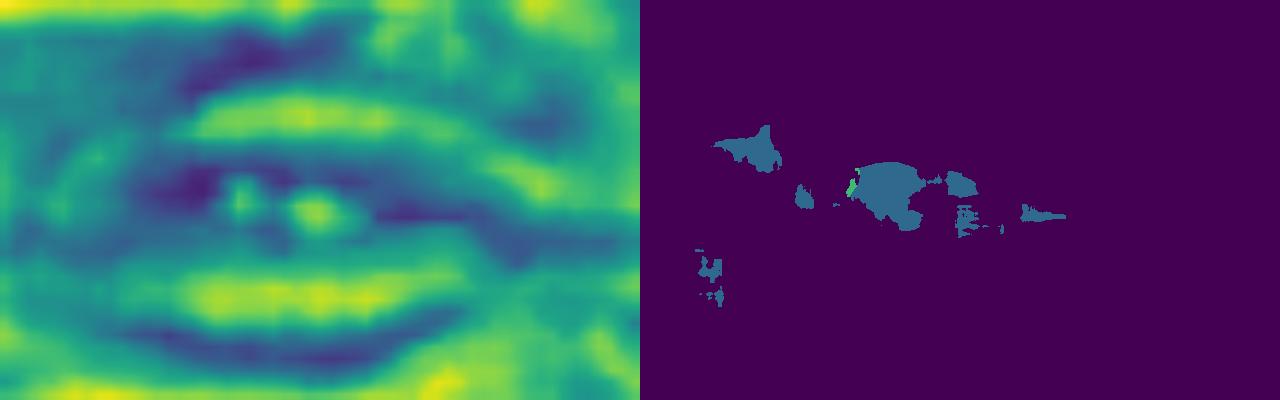}
         \caption{Crop from a face image from LFW \cite{LFWTech}}
         \label{fig:seg-ritnet-2}
     \end{subfigure}
    \caption{Eye segmentation results of RITnet \cite{chaudhary2019ritnet}}
    \label{fig:seg-ritnet}
\end{figure}

Figure \ref{fig:seg-ritnet} shows eye segmentation results of the RITnet model. In figure \ref{fig:seg-ritnet-1} it can be seen that the model achieves very good results on a high-quality image from the OpenEDS dataset. However in figure \ref{fig:seg-ritnet-2}, it can be seen that the segmentation process fails when used on an eye region crop taken from a face image of 224x224 pixels from the LFW dataset \cite{LFWTech}.

\section{Sclera Segmentation}

\begin{figure}
     \centering
     \begin{subfigure}[b]{0.24\textwidth}
         \centering
         \includegraphics[width=\textwidth]{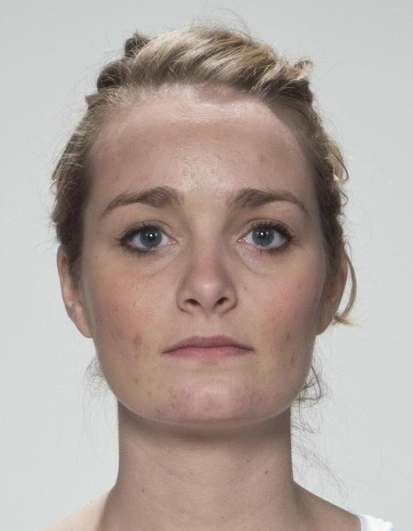}
         \caption{}
         \label{fig:seg-method-1}
     \end{subfigure}
     \hfill
     \begin{subfigure}[b]{0.24\textwidth}
         \centering
         \includegraphics[width=\textwidth]{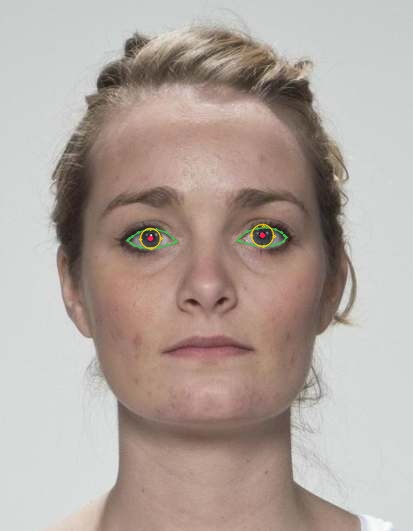}
         \caption{}
         \label{fig:seg-method-2}
     \end{subfigure}
     \hfill
     \begin{subfigure}[b]{0.24\textwidth}
         \centering
         \includegraphics[width=\textwidth]{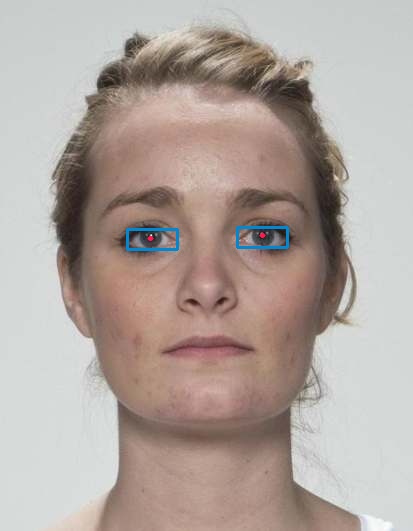}
         \caption{}
         \label{fig:seg-method-3}
     \end{subfigure}
     \hfill
     \begin{subfigure}[b]{0.24\textwidth}
         \centering
         \includegraphics[width=\textwidth]{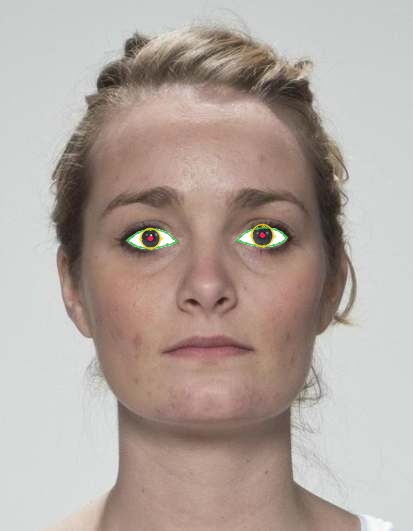}
         \caption{}
         \label{fig:seg-method-4}
     \end{subfigure}
    \caption{Landmark-based Sclera Segmentation Method. Image from FRLL \cite{DeBruine2021}}
    \label{fig:seg-method}
\end{figure}

The proposed sclera segmentation method is based on the facial landmarks and it uses MediaPipe \cite{mediapipe} as the landmarks extractor. In particular, the landmarks of both eyes as well as the landmarks of both the left and the right irises are utilized. The process is the same for both eyes so it is explained for one eye only. To decide which pixels belong to the sclera, first a convex hull of the eye's landmarks $ch(eye)$ is computed, this encloses the entire area of the eye including the sclera, the iris and the pupil. Second, the minimum enclosing circle of the iris' landmarks $ec(iris)$ is computed, this encloses the iris and the pupil. Then, all the points in the minimum bounding rectangle of the eye's landmarks $br(eye)$ that test positive for being inside the convex hull $ch(eye)$ and outside the minimum enclosing circle of the iris landmarks $ec(iris)$ (the euclidean distance between the point and the center of the circle is greater than the radius) are considered to belong to the sclera region. Figure \ref{fig:seg-method} illustrates the process, where figure \ref{fig:seg-method-1} shows the original image, figure \ref{fig:seg-method-2} shows the convex hull $ch(eye)$ in green and the enclosing circle $ec(iris)$ in yellow, figure \ref{fig:seg-method-3} shows the bounding rectangle of the eye $br(eye)$ in blue, and figure \ref{fig:seg-method-4} shows the sclera pixels painted with white.

\begin{figure}
     \centering
     \begin{subfigure}[b]{0.24\textwidth}
         \centering
         \includegraphics[width=\textwidth]{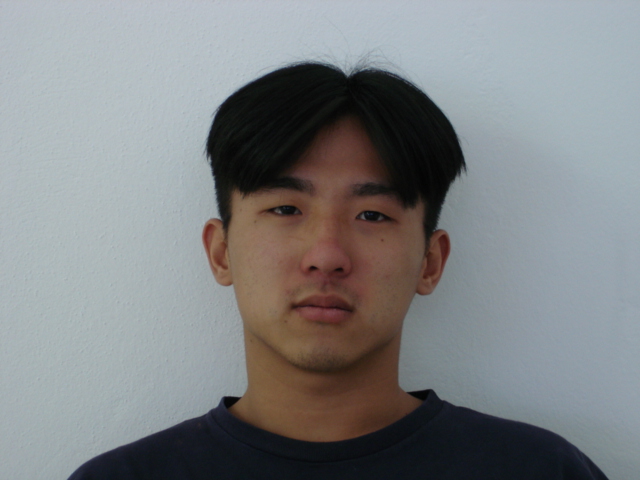}
         \label{fig:y equals x}
     \end{subfigure}
     \begin{subfigure}[b]{0.24\textwidth}
         \centering
         \includegraphics[width=\textwidth]{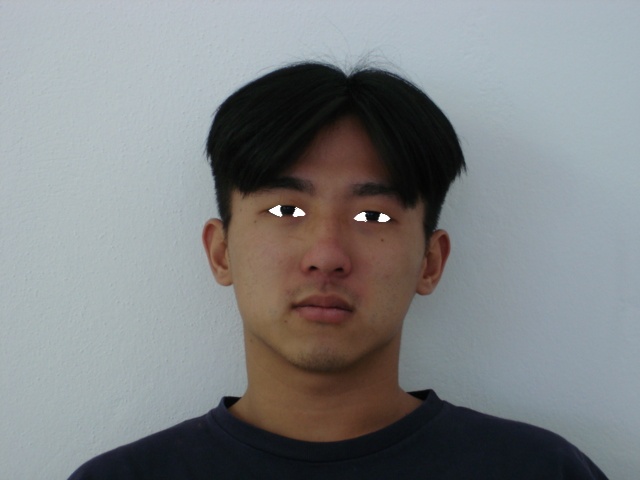}
         \label{fig:three sin x}
     \end{subfigure}
     \hfill
     \begin{subfigure}[b]{0.24\textwidth}
         \centering
         \includegraphics[width=\textwidth]{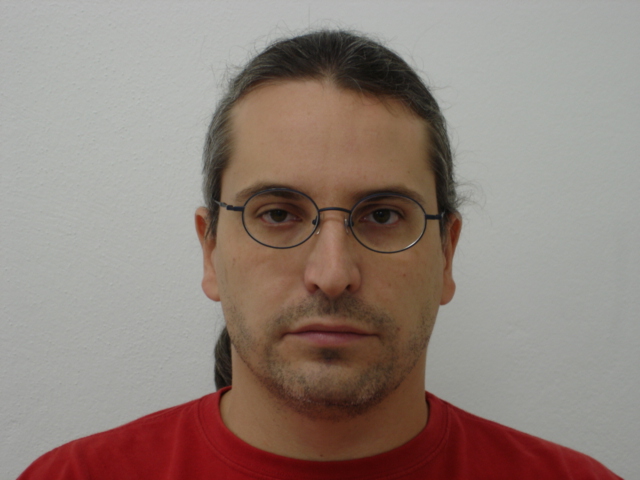}
         \label{fig:five over x}
     \end{subfigure}
     \begin{subfigure}[b]{0.24\textwidth}
         \centering
         \includegraphics[width=\textwidth]{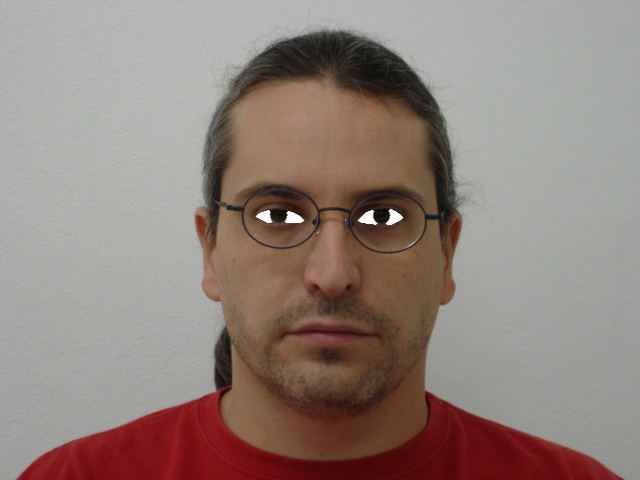}
         \label{fig:five over x}
     \end{subfigure}
     \begin{subfigure}[b]{0.24\textwidth}
         \centering
         \includegraphics[width=\textwidth]{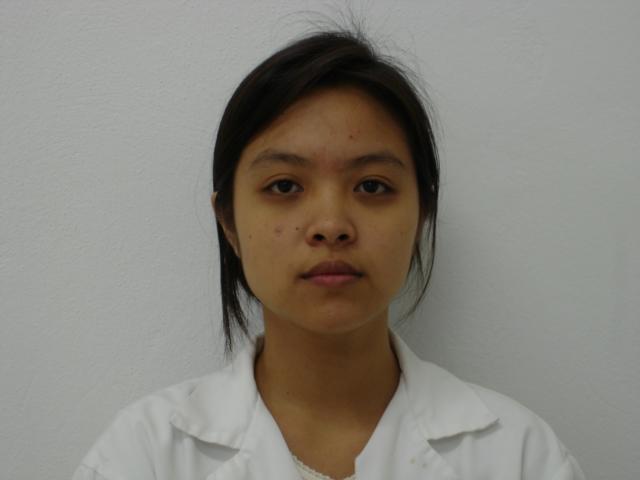}
         \label{fig:five over x}
     \end{subfigure}
     \begin{subfigure}[b]{0.24\textwidth}
         \centering
         \includegraphics[width=\textwidth]{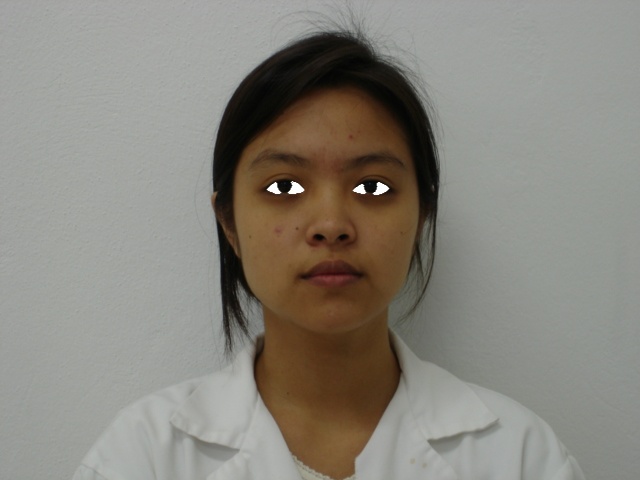}
         \label{fig:five over x}
     \end{subfigure}
     \hfill
     \begin{subfigure}[b]{0.24\textwidth}
         \centering
         \includegraphics[width=\textwidth]{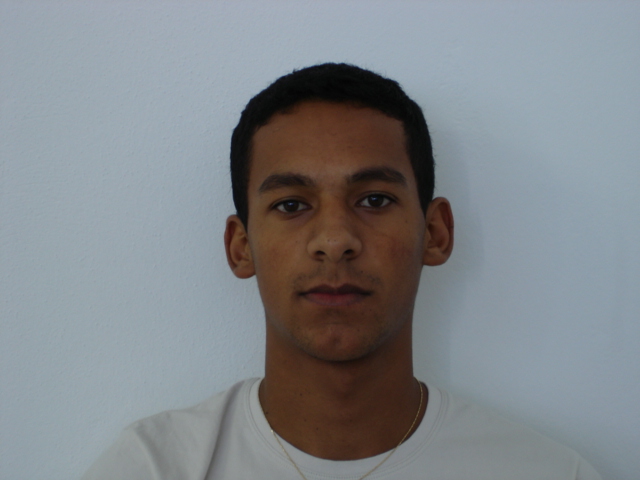}
         \label{fig:y equals x}
     \end{subfigure}
     \begin{subfigure}[b]{0.24\textwidth}
         \centering
         \includegraphics[width=\textwidth]{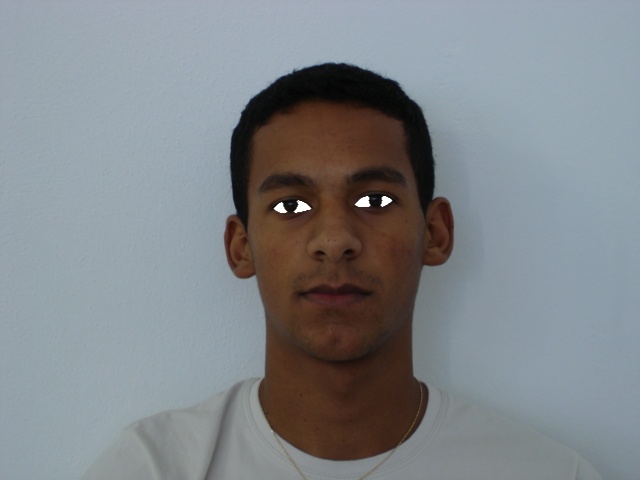}
         \label{fig:three sin x}
     \end{subfigure}
    \caption{Sclera segmentation in the enrolment scenario. Images from FEI \cite{FEI}}
    \label{fig:seg-enrolment}
\end{figure}

The landmark-based sclera segmentation method can successfully segment the sclera regardless of the skin tone, and is also robust to the size of the eyes and the presence of transparent eyeglasses, as shown in figure \ref{fig:seg-enrolment}.

\begin{figure}
     \centering
     \begin{subfigure}[b]{0.11\textwidth}
         \centering
         \includegraphics[width=\textwidth]{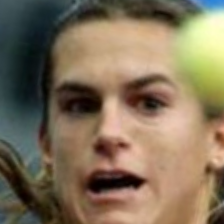}
         \label{fig:y equals x}
     \end{subfigure}
     \begin{subfigure}[b]{0.11\textwidth}
         \centering
         \includegraphics[width=\textwidth]{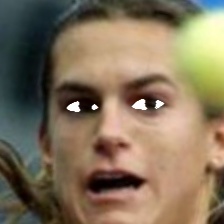}
         \label{fig:three sin x}
     \end{subfigure}
     \hfill
     \begin{subfigure}[b]{0.11\textwidth}
         \centering
         \includegraphics[width=\textwidth]{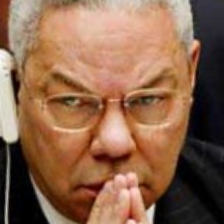}
         \label{fig:five over x}
     \end{subfigure}
     \begin{subfigure}[b]{0.11\textwidth}
         \centering
         \includegraphics[width=\textwidth]{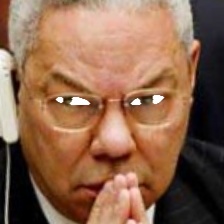}
         \label{fig:five over x}
     \end{subfigure}
     \hfill
    \begin{subfigure}[b]{0.11\textwidth}
         \centering
         \includegraphics[width=\textwidth]{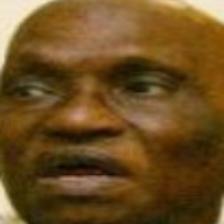}
         \label{fig:y equals x}
     \end{subfigure}
     \begin{subfigure}[b]{0.11\textwidth}
         \centering
         \includegraphics[width=\textwidth]{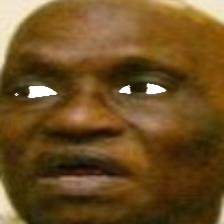}
         \label{fig:three sin x}
     \end{subfigure}
     \hfill
     \begin{subfigure}[b]{0.11\textwidth}
         \centering
         \includegraphics[width=\textwidth]{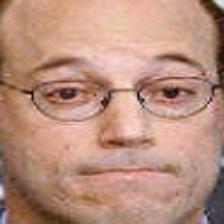}
         \label{fig:five over x}
     \end{subfigure}
     \begin{subfigure}[b]{0.11\textwidth}
         \centering
         \includegraphics[width=\textwidth]{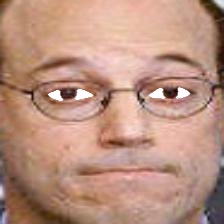}
         \label{fig:five over x}
     \end{subfigure}
     \hfill
     \begin{subfigure}[b]{0.11\textwidth}
         \centering
         \includegraphics[width=\textwidth]{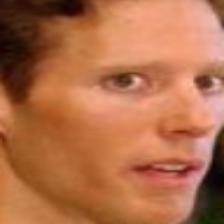}
         \label{fig:y equals x}
     \end{subfigure}
     \begin{subfigure}[b]{0.11\textwidth}
         \centering
         \includegraphics[width=\textwidth]{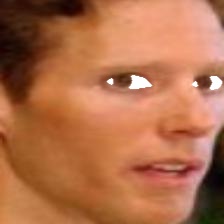}
         \label{fig:three sin x}
     \end{subfigure}
     \hfill
     \begin{subfigure}[b]{0.11\textwidth}
         \centering
         \includegraphics[width=\textwidth]{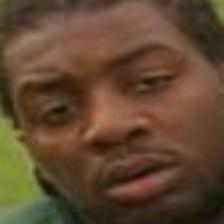}
         \label{fig:five over x}
     \end{subfigure}
     \begin{subfigure}[b]{0.11\textwidth}
         \centering
         \includegraphics[width=\textwidth]{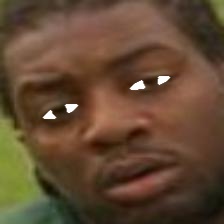}
         \label{fig:five over x}
     \end{subfigure}
     \hfill
     \begin{subfigure}[b]{0.11\textwidth}
         \centering
         \includegraphics[width=\textwidth]{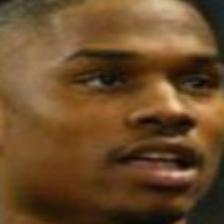}
         \label{fig:y equals x}
     \end{subfigure}
     \begin{subfigure}[b]{0.11\textwidth}
         \centering
         \includegraphics[width=\textwidth]{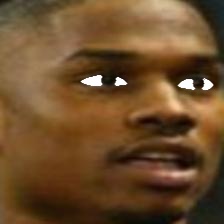}
         \label{fig:three sin x}
     \end{subfigure}
     \hfill
     \begin{subfigure}[b]{0.11\textwidth}
         \centering
         \includegraphics[width=\textwidth]{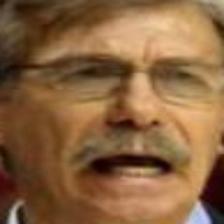}
         \label{fig:five over x}
     \end{subfigure}
     \begin{subfigure}[b]{0.11\textwidth}
         \centering
         \includegraphics[width=\textwidth]{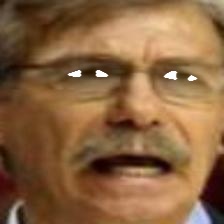}
         \label{fig:five over x}
     \end{subfigure}
    \caption{Sclera segmentation on smaller, in the wild images from LFW \cite{LFWTech}}
    \label{fig:seg-inthewild}
\end{figure}

The method works well not only in the enrolment scenario where photos of subjects are taken under controlled environment, but also works well for the border control scenario where images could be of lower quality. Figure \ref{fig:seg-inthewild} shows segmentation examples of small images of 224x224 pixels from the LFW dataset \cite{LFWTech}. The results show that the method is robust to image resolution, skin-tone, the presence of eyeglasses and small pose variations.

\section{Unnatural Color and Color Balance}

Global adjustments of color intensities that affect the entire image can result in color imbalance. Color imbalance usually takes the form of color casts or extreme color saturation. Color casts can by created synthetically as a post-processing step after taking the photo by manipulating the intensities of individual color channels. They can also result from illuminating the subject with light sources of different color temperatures, while taking the photo, causing digital cameras to render a color cast. Extreme color saturation, on the other hand, is created when the intensities of all colors in the image, not individual channels, are manipulated to take much lower or much higher values than normal resulting in under or over color saturation. 

The "No Unnatural Color" is specified as a quality component measure in ISO/IEC CD 29794-5 \cite{29794-5} because the skin color is a discriminative personal quality and thus affects the face recognition performance. However, the wide variety of skin tones and the potential presence of factors such as tattoos, moles and other facial anomalies, makes detecting unnatural color in face images reliably a challenging task.

Since the color of the sclera is uniformly whitish across all skin-tones, it should be the case that the pixel values of the sclera region show consistent changes when a face image undergoes global adjustments of color intensities such as over saturation, regardless of the skin-tone of the subject.

\begin{figure}
     \centering
     \begin{subfigure}[b]{0.19\textwidth}
         \centering
         \includegraphics[width=\textwidth]{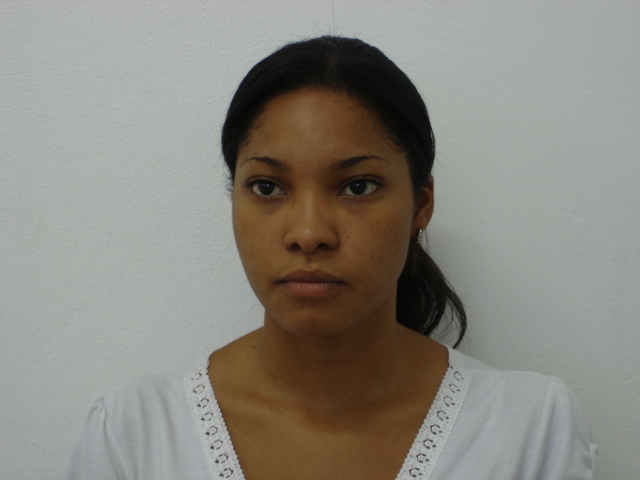}
         \caption{$s1$ Original}
         \label{fig:p1-original}
     \end{subfigure}
     \hfill
     \begin{subfigure}[b]{0.19\textwidth}
         \centering
         \includegraphics[width=\textwidth]{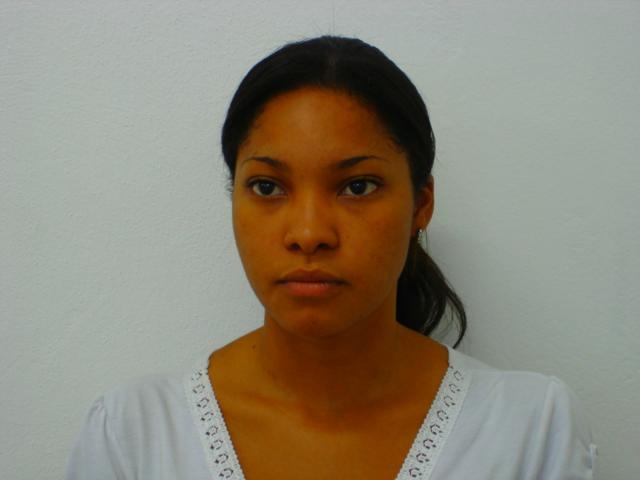}
         \caption{$f=2.0$}
         \label{fig:p1-2}
     \end{subfigure}
     \hfill
     \begin{subfigure}[b]{0.19\textwidth}
         \centering
         \includegraphics[width=\textwidth]{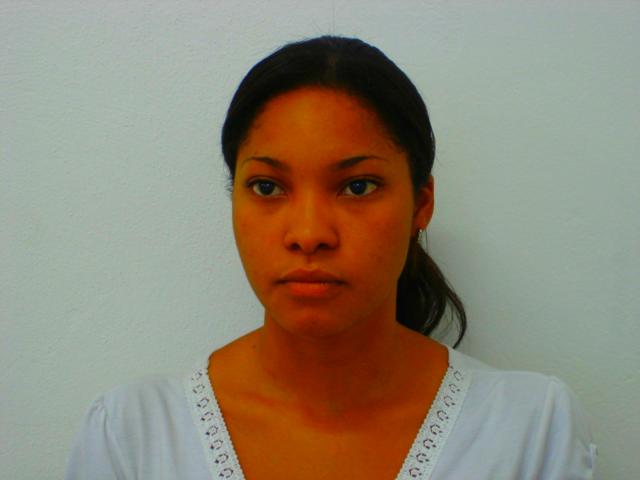}
         \caption{$f=3.0$}
         \label{fig:p1-3}
     \end{subfigure}
     \hfill
     \begin{subfigure}[b]{0.19\textwidth}
         \centering
         \includegraphics[width=\textwidth]{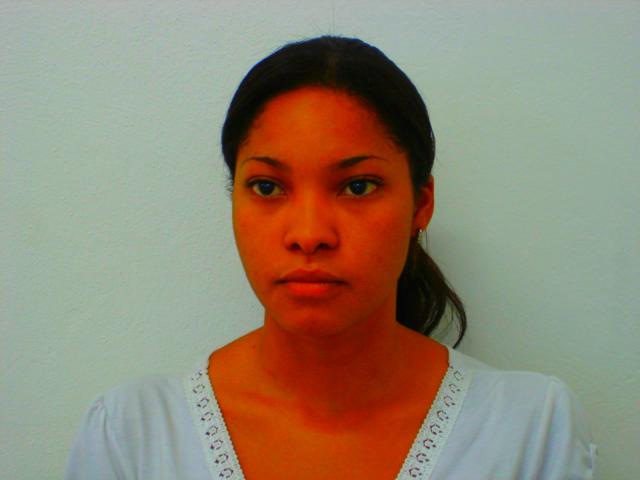}
         \caption{$f=4.0$}
         \label{fig:five over x}
     \end{subfigure}
     \hfill
     \begin{subfigure}[b]{0.19\textwidth}
         \centering
         \includegraphics[width=\textwidth]{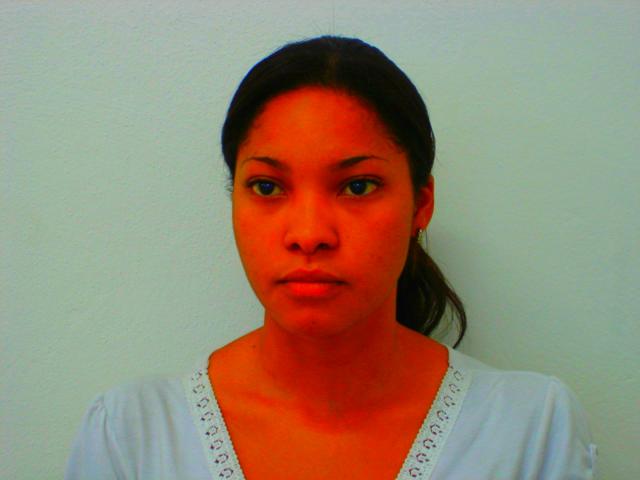}
         \caption{$f=5.0$}
         \label{fig:five over x}
     \end{subfigure}
     \begin{subfigure}[b]{0.19\textwidth}
         \centering
         \includegraphics[width=\textwidth]{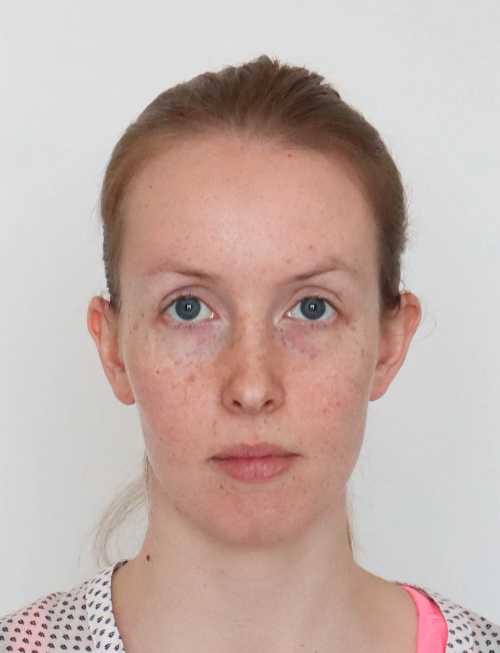}
         \caption{$s2$ Original}
         \label{fig:p2-original}
     \end{subfigure}
     \hfill
     \begin{subfigure}[b]{0.19\textwidth}
         \centering
         \includegraphics[width=\textwidth]{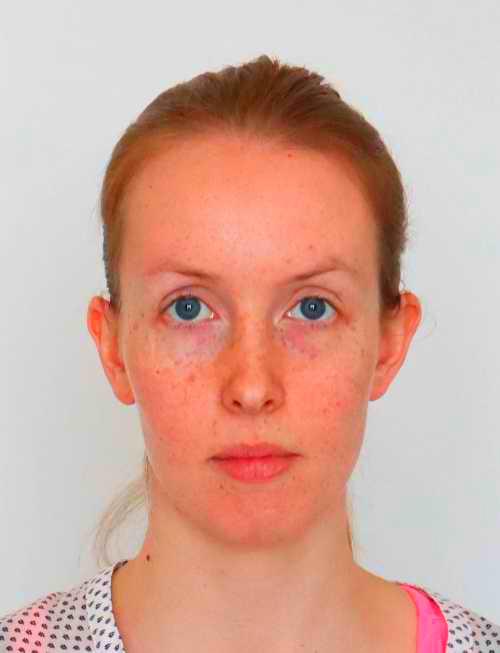}
         \caption{$f=2.0$}
         \label{fig:three sin x}
     \end{subfigure}
     \hfill
     \begin{subfigure}[b]{0.19\textwidth}
         \centering
         \includegraphics[width=\textwidth]{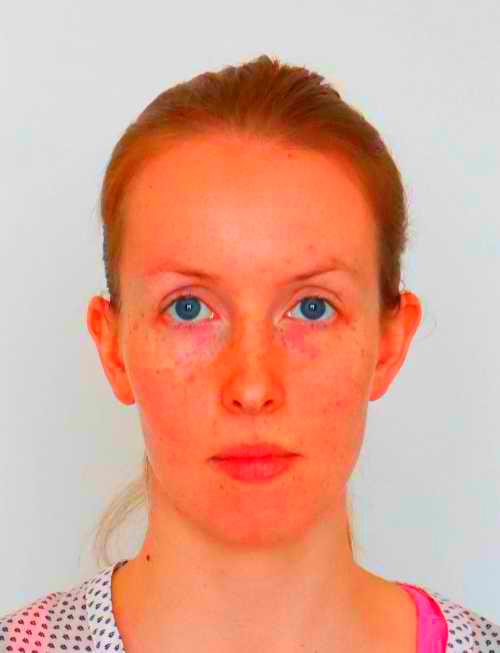}
         \caption{$f=3.0$}
         \label{fig:five over x}
     \end{subfigure}
     \hfill
     \begin{subfigure}[b]{0.19\textwidth}
         \centering
         \includegraphics[width=\textwidth]{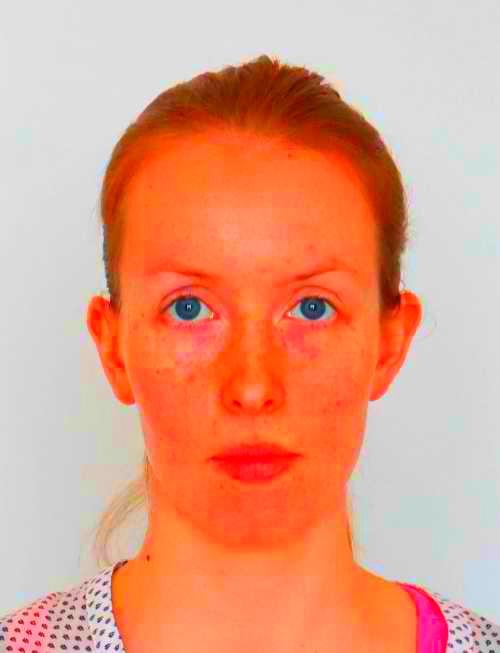}
         \caption{$f=4.0$}
         \label{fig:five over x}
     \end{subfigure}
     \hfill
     \begin{subfigure}[b]{0.19\textwidth}
         \centering
         \includegraphics[width=\textwidth]{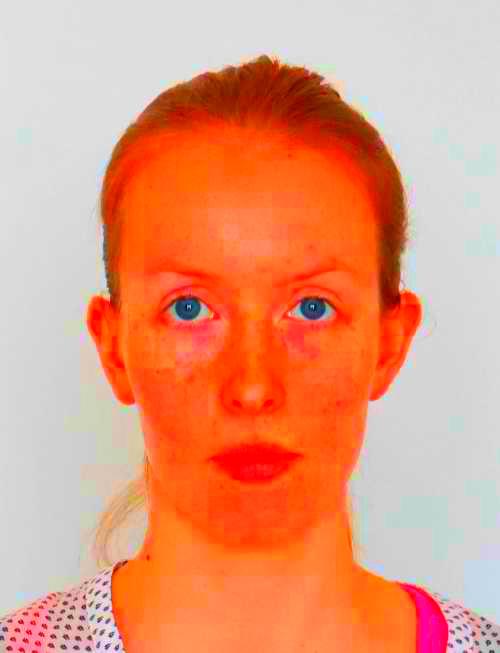}
         \caption{$f=5.0$}
         \label{fig:five over x}
     \end{subfigure}
    \caption{Images of two subjects of different skin tones with different saturation factors. Images from FEI \cite{FEI}}
    \label{fig:seg-saturation}
\end{figure}

In figure \ref{fig:seg-saturation}, sub figures \ref{fig:p1-original} and \ref{fig:p2-original} show the original images of two subjects $s1$ and $s2$ of two different skin tones. The rest of the sub figures show synthetically created images with different saturation factors, four for each of the original images. Tables \ref{tab1} and \ref{tab2} show the mean pixel values of the face region as well as the left and right sclera regions in each of the images for subject 1 and subject 2 respectively.

In table \ref{tab1}, which shows the pixel values statistics of subject $s1$ images, it can be seen that the mean pixel value of the face region increases slightly resulting in brighter color as the image gets more saturated. In table \ref{tab2}, on the other hand, which shows the statistics of subject $s2$ images, it can be seen that the mean pixel value of the face region decreases slightly resulting in darker color as the image gets more saturated. However, looking at the mean pixel values of both the left and the right sclera regions, in both tables, it can be seen that both are clearly increasing in value resulting in a brighter color in all images and for both subjects.

\begin{table}
\centering
\begin{tabular}{l l l l l l l}
\hline
 Saturation & Face Oval (s1) & Left Sclera (s1) & Right Sclera (s1)\\
\hline
Original & 60.30 & 67.28 & 71.94 \\
f=2.0 & 56.54 & 77.02 & 80.67 \\
f=3.0 & 60.96 & 83.81 & 86.97 \\
f=4.0 & 66.76 & 94.07 & 103.23 \\
f=5.0 & 72.38 & 97.45 & 106.73 \\
\hline
\end{tabular}
\caption{Mean pixel values for $s1$ images}
\label{tab1}
\end{table}

\begin{table}
\centering
\begin{tabular}{l l l l l l l}
\hline
 Saturation & Face Oval (s2) & Left Sclera (s2) & Right Sclera (s2) \\
\hline
Original & 153.60 & 165.41 & 168.16 \\
f=2.0 & 152.26 & 178.93 & 182.13 \\
f=3.0 & 145.95 & 191.13 & 192.40 \\
f=4.0 & 134.54 & 197.93 & 199.21 \\
f=5.0 & 122.48 & 212.34 & 214.46 \\
\hline
\end{tabular}
\caption{Mean pixel values for $s2$ image}
\label{tab2}
\end{table}

The purpose here is not to show an unbalanced color detection algorithm, but rather to show that a detection algorithm that relies on analyzing the sclera region, rather than the entire face, has better chances of being more reliable and skin-tone invariant. The consistent behavior of statistical values, even simple ones like the mean, across images of people with different skin colors and even different initial illumination conditions as in images \ref{fig:p1-original} and \ref{fig:p2-original}, when exposed to various color manipulations, and given that the ground truth color of the sclera region is the same, makes the detection algorithms more robust and more agnostic to demographic factors.

\section{Illumination}

In the same way in which the pixel values of the left and the right sclera regions can be used with algorithms that detect unnatural color in images, they can also be employed to get useful information about the illumination of the face images and to estimate how well the subject in a face image is illuminated and how uniform the illumination is in a way that is completely agnostic to the skin-tone of the subject.

\begin{figure}
     \centering
     \begin{subfigure}[b]{0.24\textwidth}
         \centering
         \includegraphics[width=\textwidth]{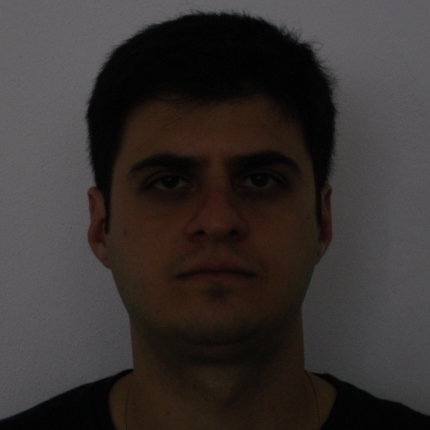}
         \includegraphics[width=\textwidth]{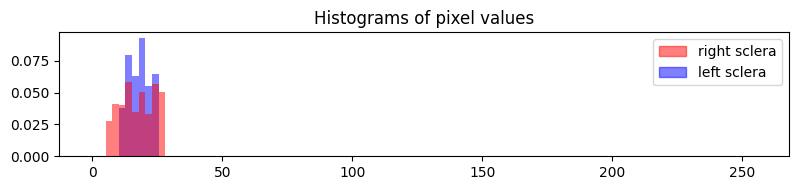}
         \caption{Bad Illumination}
         \label{fig:ill-dark}
     \end{subfigure}
     \hfill
     \begin{subfigure}[b]{0.24\textwidth}
         \centering
         \includegraphics[width=\textwidth]{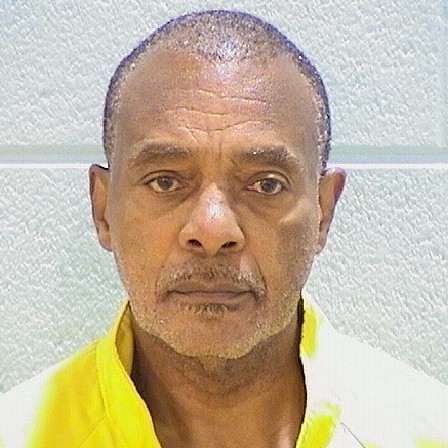}
         \includegraphics[width=\textwidth]{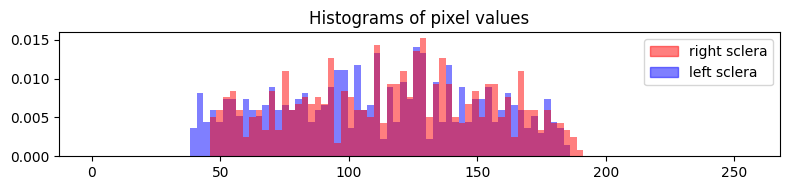}
         \caption{Good Illumination}
         \label{fig:ill-good}
     \end{subfigure}
     \begin{subfigure}[b]{0.24\textwidth}
         \centering
         \includegraphics[width=\textwidth]{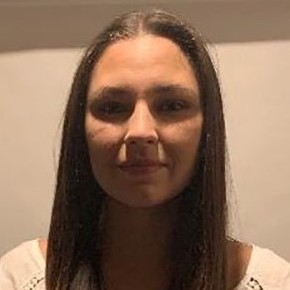}
         \includegraphics[width=\textwidth]{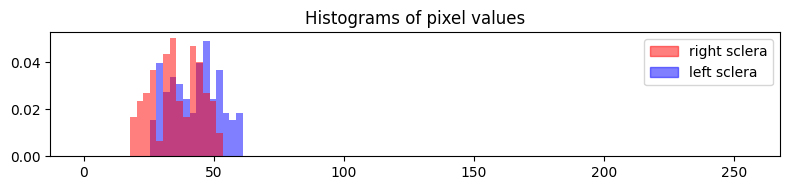}
         \caption{Top Illumination}
         \label{fig:ill-top}
     \end{subfigure}
     \hfill
     \begin{subfigure}[b]{0.24\textwidth}
         \centering
         \includegraphics[width=\textwidth]{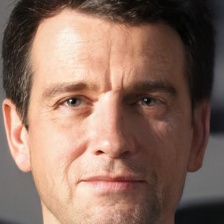}
         \includegraphics[width=\textwidth]{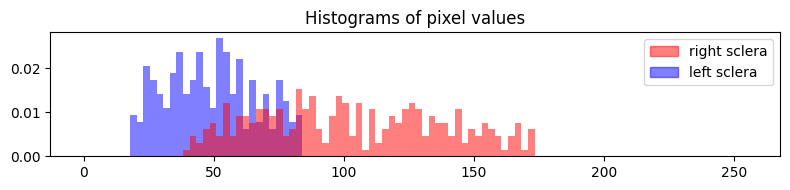}
         \caption{Nonuniform}
         \label{fig:ill-nonuni}
     \end{subfigure}
    \caption{Sclera segmentation for assessing face image illumination. Images from FEI \cite{FEI}, LFW \cite{LFWTech}, and Illinois DOC labeled faces \cite{Illionis}}
    \label{fig:illumination}
\end{figure}

Figure \ref{fig:illumination} shows examples of face images with varying illumination quality. Figure \ref{fig:ill-dark} shows a very dark image where the subject is barely visible. Looking at the histogram of the pixel values of the sclera regions, it can be clearly seen that most pixel values are on the lower end of the value scale and thus have darker colors. Figure \ref{fig:ill-good} shows an image with well illuminated subject. This can also be deduced by looking at the histogram which also confirms that the illumination is symmetric between the left and the right side. Figure \ref{fig:ill-top} shows a poorly lit subject, with a light source from the top causing the eyes region to be dark. The same information can also be deduced by looking at the histogram of the pixel values of the sclera region. Lastly, figure \ref{fig:ill-nonuni} shows a face image with non-uniform illumination, where the right side of the face is well illuminated while the left side is rather dark. This can be confirmed by looking at the histograms which show that the distribution of the pixel values of the left sclera is shifted to the left and has lower pixel values, thus the left side is less illuminated than the right side of the face. All this analysis can be done independently from the subject in the image and without considering what skin-tone they might have.

\section{Conclusion and Future Work}

Face image quality assessment algorithms should be invariant to demographic factors. The eye sclera is one region of the face which has consistent whitish color across demographic boundaries. This work introduced a novel algorithm for sclera segmentation that is suitable for face images used during the enrolment and the verification scenarios. It then presented how the behavior of the statistical features of the pixel values of the sclera regions is consistent across different demographic boundaries which makes them very useful for creating FIQA algorithms that are more robust and invariant to demographic factors \footnote{https://github.com/wkabbani/sclera-segmentation}.

A follow up work will utilize the sclera segmentation method and incorporate the demonstrated consistent statistical behavior of the pixel values of the sclera regions into the face image quality  assessment algorithms of various face image quality components to make them more robust to demographic factors.

\section{Acknowledgement}
This work was supported by the European Union’s Horizon 2020 Research and Innovation Program under Grant 883356.


\begin{thebibliography}{Wa20}

\bibitem[Ch19]{chaudhary2019ritnet}
Chaudhary, Aayush~K; Kothari, Rakshit; Acharya, Manoj; Dangi, Shusil; Nair, Nitinraj; Bailey, Reynold; Kanan, Christopher; Diaz, Gabriel; Pelz, Jeff~B: , RITnet: real-time semantic segmentation of the eye for gaze tracking, 2019.

\bibitem[DJ21]{DeBruine2021}
DeBruine, Lisa; Jones, Benedict: , {Face Research Lab London Set}, 4 2021.

\bibitem[Fi19]{Illionis}
Fisher, David~J.: , Illinois DOC labeled faces dataset, 2019.
\newblock \url{https://www.kaggle.com/davidjfisher/illinois-doc-labeled-faces-dataset}.

\bibitem[Ga19]{OpenEDS}
Garbin, Stephan~J.; Shen, Yiru; Schuetz, Immo; Cavin, Robert; Hughes, Gregory; Talathi, Sachin~S.: , OpenEDS: Open Eye Dataset, 2019.

\bibitem[Hu07]{LFWTech}
Huang, Gary~B.; Ramesh, Manu; Berg, Tamara; Learned-Miller, Erik: , Labeled Faces in the Wild: A Database for Studying Face Recognition in Unconstrained Environments, October 2007.

\bibitem[IS]{29794-5}
ISO/IEC WD 29794-5 Information technology — Biometric sample quality — Part 5: Face image data.
\newblock \url{https://www.iso.org/standard/81005.html}.

\bibitem[Ka23]{KANO202310}
Kano, Fumihiro: , Evolution of the Uniformly White Sclera in Humans: Critical Updates, 2023.

\bibitem[Ki19]{NVGaze}
Kim, Joohwan; Stengel, Michael; Majercik, Alexander; Mello, Shalini; Dunn, David; Laine, Samuli; McGuire, Morgan; Luebke, David: , NVGaze: An Anatomically-Informed Dataset for Low-Latency, Near-Eye Gaze Estimation, 05 2019.

\bibitem[Lu19]{mediapipe}
Lugaresi, Camillo; Tang, Jiuqiang; Nash, Hadon; McClanahan, Chris; Uboweja, Esha; Hays, Michael; Zhang, Fan; Chang, Chuo-Ling; Yong, Ming; Lee, Juhyun; Chang, Wan-Teh; Hua, Wei; Georg, Manfred; Grundmann, Matthias: , MediaPipe: A Framework for Perceiving and Processing Reality, 2019.

\bibitem[Th]{FEI}
Thomaz, Carlos~Eduardo: , FEI Face Database.
\newblock \url{https://fei.edu.br/~cet/facedatabase.html}.

\bibitem[Wa19]{wang2019sclerasegnet}
Wang, Caiyong; He, Yong; Liu, Yunfan; He, Zhaofeng; He, Ran; Sun, Zhenan: , ScleraSegNet: an Improved U-Net Model with Attention for Accurate Sclera Segmentation, 2019.

\bibitem[Wa20]{wang2020sclerasegnet}
Wang, Caiyong; Wang, Yunlong; Liu, Yunfan; He, Zhaofeng; He, Ran; Sun, Zhenan: , ScleraSegNet: an Attention Assisted U-Net Model for Accurate Sclera Segmentation, 2020.

\bibitem[Zh12]{sclreg}
Zhou, Zhi; Du, Eliza~Yingzi; Thomas, N.~Luke; Delp, Edward~J.: , A New Human Identification Method: Sclera Recognition, 2012.

\end{thebibliography}
\end{document}